\documentclass{ieeeaccess}
\usepackage{cite}
\usepackage{amsmath,amssymb,amsfonts}
\usepackage{algorithmic}
\usepackage{algorithm}
\usepackage{graphicx}
\usepackage{textcomp}

\usepackage{booktabs}
\usepackage{multirow}
\usepackage{soul}

\usepackage{bm}
\makeatletter
\AtBeginDocument{\DeclareMathVersion{bold}
\SetSymbolFont{operators}{bold}{T1}{times}{b}{n}
\SetSymbolFont{NewLetters}{bold}{T1}{times}{b}{it}
\SetMathAlphabet{\mathrm}{bold}{T1}{times}{b}{n}
\SetMathAlphabet{\mathit}{bold}{T1}{times}{b}{it}
\SetMathAlphabet{\mathbf}{bold}{T1}{times}{b}{n}
\SetMathAlphabet{\mathtt}{bold}{OT1}{pcr}{b}{n}
\SetSymbolFont{symbols}{bold}{OMS}{cmsy}{b}{n}
\renewcommand\boldmath{\@nomath\boldmath\mathversion{bold}}}
\makeatother

\def\BibTeX{{\rm B\kern-.05em{\sc i\kern-.025em b}\kern-.08em
    T\kern-.1667em\lower.7ex\hbox{E}\kern-.125emX}}

\begin{document}
\history{Date of publication 27 May, 2026, date of current version 1 June, 2026.}
\doi{10.1109/ACCESS.2026.3697689}

\title{ClustRecNet: A Novel End-to-End Deep Learning Framework for Clustering Algorithm Recommendation}
\author{\uppercase{Mohammadreza Bakhtyari}\authorrefmark{1},
\uppercase{Bogdan Mazoure}\authorrefmark{2}, \uppercase{Renato Cordeiro de Amorim}\authorrefmark{3} \uppercase{Guillaume Rabusseau}\authorrefmark{2,4}, \uppercase{Vladimir Makarenkov}\authorrefmark{1,2}}

\address[1]{Département d'Informatique, Université du Québec à Montréal, Montréal, QC, H2X 3Y7, Canada}
\address[2]{Mila - Quebec AI Institute, Montréal, QC, H2S 3H1, Canada}
\address[3]{School of Computer Science and EE, University of Essex, Colchester, CO4 3SQ, UK}
\address[4]{Department of Computer Science and Operations Research, Université de Montréal, Montréal, QC, H3T 1N8, Canada}

\tfootnote{This work was supported by le Fonds Québécois de la Recherche sur la Nature et les Technologies [grant 371537] and the Natural Sciences and Engineering Research Council of Canada [grant 249644].}

% \markboth
% {Bakhtyari \headeretal: Preparation of Papers for IEEE TRANSACTIONS and JOURNALS}
% {Bakhtyari \headeretal: Preparation of Papers for IEEE TRANSACTIONS and JOURNALS}

\corresp{Corresponding author: Vladimir Makarenkov (e-mail: makarenkov.vladimir@uqam.ca).}

\begin{abstract}
Identifying an effective clustering algorithm for a given dataset remains a fundamental unsupervised learning issue. We introduce ClustRecNet, a novel end-to-end deep learning framework that recommends suitable clustering algorithm(s) by directly learning high-order representations of raw tabular data. To facilitate robust meta-learning, we first construct a comprehensive repository of $34,000$ synthetic datasets encompassing a large variety of clustering scenarios, run 10 popular clustering algorithms, and use Adjusted Rand Index (ARI) to establish ground-truth labels. ClustRecNet's architecture incorporates a convolution block, two residual blocks, and an attention block to capture local and global structural patterns, effectively bypassing the knowledge bottleneck associated with manual feature engineering. Extensive evaluation on both synthetic and real-world benchmarks demonstrates that ClustRecNet consistently outperforms traditional internal cluster validity indices such as Silhouette, Calinski-Harabasz, Davies–Bouldin, and Dunn as well as state-of-the-art Automated Machine Learning (AutoML) approaches such as ML2DAC, AutoCluster, and AutoML4Clust. For example, our framework achieves an average $0.497$ ARI gain over the Calinski–Harabasz cluster validity index on synthetic data and an average $44.16\%$ ARI improvement over the leading AutoML approach (ML2DAC) on real-world benchmarks. Code and data are available at: https://github.com/mrbakhtyari/ClustRecNet  
\end{abstract}


\titlepgskip=-21pt

\maketitle

\section{Introduction} \label{intro}
Clustering is a cornerstone of unsupervised learning, enabling the extraction of patterns from unlabeled data in complex domains such as medical informatics, social network analysis, and anomaly detection \cite{mirkin2005clustering,ezugwu2022app,oyewole2023data,zhou2024survey}. However, the utility of the extracted knowledge is strictly influenced by the selection of an appropriate clustering algorithm. Despite the proliferation of diverse clustering methodologies – ranging from classical partitioning \cite{macqueen67} to recent deep learning clustering\cite{xie2016dec}, and adaptive hybrid frameworks \cite{abhishek2026adaptive} – each characterized by specific inductive biases and operational constraints, identifying a suitable clustering algorithm for a given dataset remains a non-trivial challenge. While existing comparative frameworks \cite{brock08, arbelaitz13, Rodriguez19, wani24} offer valuable insights, the lack of an automated, data-driven mechanism for clustering algorithm selection remains a significant bottleneck in the development of fully autonomous intelligent systems.

This challenge is further compounded by the difficulty of objectively validating clustering outcomes in the absence of prior labels, necessitating considerable domain knowledge and extensive trial-and-error experimentation. A crucial step in this process involves evaluating the quality of clustering solutions, which is typically carried out using established cluster validity indices (CVIs) \cite{HASSAN24, de2016applying,rykov2024inertia}. The indices such as Silhouette (Sil) \cite{Rousseeuw87}, Calinski-Harabasz (CH) \cite{Calinski74}, Davies–Bouldin (DB) \cite{davies79}, and Dunn \cite{dunn73} reflect different criteria of cluster compactness and separation. However, these CVIs often fall short when dealing with datasets characterized by complex structures, outliers, or overlapping clusters, thereby revealing critical limitations in existing cluster validation methodologies \cite{arbelaitz13, rykov2024inertia}.

Recently, several Automated Machine Learning (AutoML) approaches have been introduced to address the issue of selecting the most suitable clustering algorithm for a given dataset in unsupervised settings \cite{Francia24,dilmperis2025dataset2graph}. Many of them are inspired by established AutoML techniques in supervised learning, particularly those tackling the Combined Algorithm Selection and Hyperparameter optimization (CASH) problem \cite{thornton13}. The literature predominantly outlines two main strategies. The first of them jointly optimizes the choice of clustering algorithm and its hyperparameter space, as exemplified by AutoML4Clust \cite{tschechlov21}. However, such an approach is often computationally expensive, particularly on large and complex search spaces. The second, more prevalent, strategy treats the CASH problem sequentially — first selecting a suitable clustering algorithm, followed by hyperparameter tuning. The approaches supporting this strategy leverage meta-learning techniques that transfer knowledge from existing datasets.

Early work by De Souto et al. \cite{desouto08} pioneered the use of meta-learning for selecting the most appropriate clustering algorithm(s) for a given dataset, thus laying the groundwork for subsequent research in this domain. Most AutoML systems for clustering follow a two-stage architecture: (1) an offline phase that extracts manual meta-features and associates them with algorithm performance; (2) an online phase that employs shallow learners such as $k$-nearest neighbors or decision trees to recommend the most convenient clustering algorithms for new datasets. For example, AutoCluster \cite{liu21} leverages landmark-based and statistical meta-features and subsequently applies a grid search strategy
for configuration optimization. In contrast, ML2DAC \cite{treder23} relies on a diversified set of data descriptors, including general properties (e.g. instance counts), statistical moments, and information-theoretic measures, to recommend both clustering algorithms and their optimal hyperparameters. 

Despite their key strengths, existing meta-learning-based methods face several inherent limitations. The reliance on manually engineered meta-feature vectors introduces a significant representation gap; these features typically act as lossy compression of high-dimensional data into global statistics, thereby obscuring the local manifold geometry and non-linear dependencies that are fundamental to cluster separability. Consequently, such descriptors often fail to capture the structural knowledge required to distinguish between clustering algorithms with different inductive biases.
% that may obscure essential structural knowledge and distributional characteristics of the data, which critically influence clustering behavior and algorithm performance. 
Moreover, the selection of informative and representative meta-features remains an open problem, with no established consensus on optimal descriptors \cite{poulakis24}. Finally, the prevailing modular architecture – separating feature extraction from model training – imposes an information bottleneck. By decoupling these phases, the system cannot adaptively refine its internal representations in a task-specific manner, which limits the expressiveness and adaptability of learned representations, potentially constraining generalization to complex datasets.

In this paper, we propose ClustRecNet, a novel deep learning framework that eliminates the need for handcrafted meta-feature engineering and facilitates the automated discovery of data geometry to recommend suitable clustering algorithms for tabular data at hand. The main contributions of our work are as follows:

\begin{itemize}
    \item We introduce a deep learning architecture that operates directly on input data distributions, bypassing the manual engineering of meta-features and thus preserving the local manifold geometry that traditional global statistics typically obscure.
    % that learns latent structural representations directly from input data distributions to recommend suitable clustering algorithms.
    \item We eliminate the need for subjective selection of meta-feature descriptors by enabling the model to automatically learn and prioritize informative representations, thereby resolving the lack of consensus on optimal meta-features.
    \item With an end-to-end architecture that enables a joint optimization of feature discovery and algorithm recommendation, unlike modular pipelines, our framework allows the recommendation loss to guide the feature extraction process through backpropagation, ensuring that the discovered latent representations are specifically refined to maximize clustering algorithm recommendation accuracy.
    % We eliminate the information bottleneck caused by manual meta-feature engineering, allowing the model to capture high-order structural patterns.
    \item We carry out an extensive empirical evaluation of the new model on both synthetic and real-world benchmarks, demonstrating its superior generalization capability compared to three state-of-the-art AutoML methods (AutoCluster, AML4C, and ML2DAC) and four traditional cluster validity indices (Silhouette, Calinski-Harabasz, Davies–Bouldin, and Dunn).
\end{itemize}

The paper is organized as follows: Section 2 presents the data generation protocol, clustering algorithms being used, the proposed ClustRecNet architecture, and the training process. Section 3 describes the experimental setup and reports results from experiments with synthetic and real data. Finally, Section 4 concludes the study, summarizing its key findings and implications.

\section{Methodology}\label{sec:method}

The proposed recommendation system introduces a novel deep learning framework for selecting clustering algorithms, which overcomes key limitations of traditional validation indices and recent AutoML approaches. Our method employs a network architecture trained on a diverse repository of 34,000 datasets, enabling it to learn structural patterns across varying clustering scenarios. To generate training labels, we evaluated the performance of 10 widely-used clustering algorithms on each dataset considered, using the Adjusted Rand Index (ARI) \cite{Hubert85} (i.e. the best-performing algorithm(s) correspond to the highest ARI value(s)). Unlike internal CVIs, which are often constrained by specific geometric assumptions regarding cluster compactness or separation, ARI provides an objective, ground-truth-based evaluation of an algorithm's ability to recover the true underlying structure, serving as an unbiased supervisor for our meta-learning framework.
Once trained on these ARI-labeled datasets, our model can effectively recommend the most suitable clustering algorithms for previously unseen data. As we will see, such an ARI-based training not only automates the clustering algorithm selection process but also improves clustering accuracy and generalization, eliminating the need for handcrafted meta-features and CVIs.

This section provides a detailed exposition of the methodology used to develop the ClustRecNet framework, the high-level logic of which is summarized in Algorithm \ref{alg:clustrecnet}. Here, we describe the synthetic data used for training and validation, outline the clustering algorithms evaluated on these data, and present details of the proposed network architecture and the training procedure used to optimize its performance.

\begin{algorithm}[h]
\caption{ClustRecNet}
\label{alg:clustrecnet}
\begin{algorithmic}[1]
\STATE \textbf{Phase I: Data Generation \& Labeling (Offline)}
\FOR{each configuration $c \in \{Scenario 1, Scenario 2\}$}
    \STATE Generate $2,000$ synthetic datasets $X_{synthetic}$ using controlled parameters $\{K, N, D, \alpha, \dots\}$
\ENDFOR
\FOR{each dataset $D_i \in \{34,000$ datasets$\}$}
    \FOR{each clustering algorithm $A_j \in \{A_1, \dots, A_{10}\}$}
        \STATE $Solution_{ij} \leftarrow$ Apply algorithm $A_j$ to $D_i$
        \STATE $Score_{ij} \leftarrow$ Calculate $ARI(Solution_{ij}, GroundTruth_i)$
        \STATE $y_{ij} \leftarrow 1$ \textbf{if} $Score_{ij} \ge 0.8$ \textbf{else} $0$
    \ENDFOR
    \STATE Label vector $Y_i \leftarrow [y_{i1}, y_{i2}, \dots, y_{i10}]$
\ENDFOR

\STATE \textbf{Phase II: End-to-End Model Training (Offline)}
\STATE Preprocess all $D_i$ (symmetric padding)
\STATE Initialize ClustRecNet architecture $\Theta$
\FOR{epoch $= 1$ to $30$}
    \STATE $\hat{Y} \leftarrow$ Forward pass ClustRecNet($D_i, \Theta$)
    \STATE $\mathcal{L} \leftarrow$ Calculate Multi-label BCE Loss($\hat{Y}, Y_i$)
    \STATE $\Theta \leftarrow$ Update weights via Adam optimizer
\ENDFOR

\STATE \textbf{Phase III: Recommendation (Online)}
\STATE \textbf{Input:} Unseen tabular dataset $D_{new}$
\STATE $D'_{new} \leftarrow$ Preprocess $D_{new}$
\STATE $Probs \leftarrow$ Forward pass ClustRecNet($D'_{new}, \Theta$)
\STATE \textbf{Output:} Recommended clustering algorithm(s)

\end{algorithmic}
\end{algorithm}

\subsection{Data Generation} \label{sec:data_gen}

To train a robust DL (deep learning)-based recommendation framework, we generated synthetic datasets with controlled variation across key parameters, including the number and shape of clusters, the number of objects and features, intra-cluster density, degree of cluster overlap, and level of noise added to feature values. Accounting for these factors ensures that the model can generalize across a wide range of clustering scenarios and provide accurate clustering algorithm recommendations. 

Precisely, synthetic data were generated under two distinct scenarios:

1) Scenario 1 follows the data generation strategy proposed by Rodriguez et al. \cite{Rodriguez19}. It relies on five key parameters: \(K\) - the number of clusters in the dataset, $N$ - the number of objects in the dataset, \(D\) - the number of features (i.e. number of available dimensions) for each object, \(N_e\) - the number of objects per cluster, and \(\alpha\) - the cluster separation parameter (i.e. degree of overlap between clusters). To cover a broad range of clustering complexities in our simulations, we varied the number of clusters, features, and objects per cluster. Specifically, we considered the values of 2, 5, and 10 for the number of clusters; 2, 5, 10, and 50 for the number of features; and 50 and 200 for the number of objects per cluster. This corresponds to 24 different parameter configurations.  To select appropriate values of the \(\alpha\) parameter, which controls inter-cluster separation and depends on both the number of clusters and dimensionality, we performed a grid search over the range [0.1, 10] with a step size of 0.1. The selected \(\alpha\) values satisfied two criteria: (1) no single clustering algorithm consistently achieved an ARI above 0.8 across all datasets of any given parameter configuration, and (2) at least one algorithm exceeded this threshold in more than 20\% of the realizations. These constraints prevented any single algorithm from clearly dominating the others. In total, we identified 11 parameter configurations meeting these criteria.

2) Scenario 2 corresponds to an alternative data generation strategy proposed by Zellinger and Buhlmann \cite{zellinger23}. We used this strategy when no values of \(\alpha\) were satisfying the desired ARI conditions for a given parameter configuration. Instead of considering the cluster separation parameter \(\alpha\), the procedure of Zellinger and Buhlmann relies on a richer set of parameters, including cluster overlap, aspect ratio, radius ratio, cluster distribution, and imbalance ratio. In these settings, we explored three different minimum-maximum overlap intervals: \((0.001, 0.002)\), \((0.001, 0.25)\), and \((0.25, 0.30)\). As the aspect ratio (i.e. the ratio between the longest and the shortest axes in a cluster), we used the values of 1 and 3. As the radius ratio (i.e. the ratio between the radii of the largest and the smallest clusters), we used the values of 1, 3, and 10. Cluster distributions considered included normal, exponential, and $t$ distributions. Finally, the cluster imbalance ratios of 1, 3, and 5 were explored. To mitigate bias, we calibrated parameter combinations under the %identical 
same conditions as in Scenario 1.
As a result, we identified 6 different valid parameter configurations for this scenario.

\begin{figure}[t]
\centering
\includegraphics[width=\columnwidth]{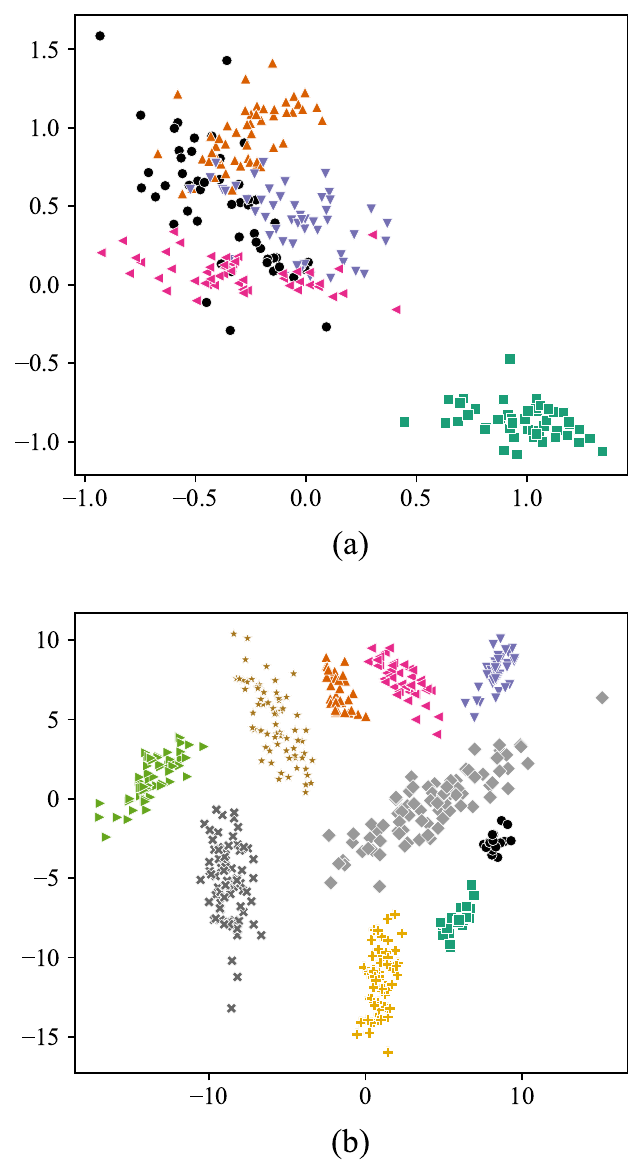}
\caption{Scatter plots illustrating the distribution of objects in two sample datasets: (a, Scenario 1) Clusters generated according to the strategy of Rodriguez et al. \cite{Rodriguez19}, using the following parameters: $K=5$, $N=1000$, $D=2$, $N_e=200$, and $\alpha=5$; (b, Scenario 2) Clusters generated according to the strategy of Zellinger and Buhlmann \cite{zellinger23}, using the following parameters: $K=10$, $N=500$, $D=2$, cluster overlap $(0.001, 0.002)$, aspect ratio $3$, radius ratio $3$, normal distribution, and imbalance ratio $5$.}
\label{fig:scatterplot}
\end{figure}

Overall, we generated 2,000 datasets for each of the 17 retained configurations (11 from Scenario 1 and 6 from Scenario 2) to ensure an empirical trade-off between dataset diversity and training stability, resulting in a comprehensive data repository of 34,000 synthetic datasets (available at our GitHub repository\footnote{https://github.com/mrbakhtyari/ClustRecNet}). Fig. \ref{fig:scatterplot} presents examples of datasets from Scenarios 1 and 2. To ensure a consistent input scale for training, we normalized the features in all datasets using $z$-scores. Additionally, we applied symmetric padding to align the input shapes across datasets of varying sizes.

\subsection{Clustering Algorithms}\label{subsec:clustalg}

The set of clustering algorithms considered in our study includes partitioning methods such as $k$-means \cite{macqueen67} and $k$-medians \cite{jain88}, hierarchical approaches such as Agglomerative Clustering (AC) \cite{defays77}, BIRCH \cite{zhang96}, and Ward's method \cite{ward63}, and density-based techniques such as DBSCAN \cite{ester96}, HDBSCAN \cite{campello13}, and OPTICS \cite{ankerst99}. Moreover, we used the probabilistic Gaussian Mixture Models (GMM) \cite{reynolds09} and the graph-based Spectral Clustering (SC) \cite{ng01} techniques for their unique properties and advantages. All implementations were carried out using the scikit-learn library, and where required (e.g. in $k$-means), the number of clusters was assumed to be known. Given the extensive literature on parameter selection for clustering algorithms, we did not analyze parameter effects comprehensively. Therefore, instead of optimizing hyperparameters for each realization, we performed grid search once per configuration to maximize the mean of ARI across all 2,000 datasets generated under each of our 17 parameter configurations, thus ensuring consistent and reproducible tuning while significantly reducing computational overhead. The hyperparameter selection details for the 10 clustering algorithms considered in our study are available in Section \ref{subsec:hyper}.

\subsubsection{Hyperparameters Used with Clustering Algorithms} \label{subsec:hyper}

\begin{table*}[t]
\centering
\caption{Summary of hyperparameters and search spaces used for each clustering algorithm used in our experiments.}
\label{tab:hyperparameters}
\begin{tabular}{lll}
\toprule
\textbf{Algorithm} & \textbf{Hyperparameter} & \textbf{Range / Search Values} \\
\midrule
$k$-means & \textit{n\_init} & 100 \\
$k$-medians & \textit{n\_init} & 100 \\
\midrule
Ward & \textit{n\_neighbors} & $ [3, 10]$, step 1 \\
\midrule
Agglomerative (AC) & \texttt{linkage} & $\{ \text{complete, average, single} \}$ \\
& \texttt{metric} & $\{ \text{euclidean, l1, l2, cosine} \}$ \\
& \texttt{n\_neighbors} & $[3, 10]$, step 1 \\
\midrule
DBSCAN & \texttt{eps} & $[0.1, 1.5]$, step 0.1 \\
& \texttt{min\_samples} & 5 \\
\midrule
HDBSCAN & \texttt{min\_samples} & $[3, 10]$, step 1 \\
 & \texttt{min\_cluster\_size} & $[5, 25]$, step 5 \\
\midrule
OPTICS & \texttt{min\_samples} & $[3, 10]$, step 1 \\
 & \texttt{min\_cluster\_size} & $[0.1, 0.5]$, step 0.05 \\
 & \texttt{xi} & $[0.05, 0.25]$, step 0.05 \\
\midrule
Spectral Clustering (SC) & \texttt{eigen\_solver} & $\{ \text{arpack, lobpcg} \}$ \\
 & \texttt{affinity} & $\{ \text{nearest\_neighbors, rbf} \}$ \\
\midrule
BIRCH & \texttt{threshold} & $[0.1, 1.5]$, step 0.1 \\
\midrule
GMM & \texttt{covariance\_type} & $\{ \text{full, tied, diagonal, spherical} \}$ \\
\bottomrule
\end{tabular}
\end{table*}

To ensure a fair and comprehensive comparison across clustering algorithms, we systematically explored a curated set of hyperparameter configurations for each of them, while preserving computational tractability. The selected configurations aim to balance representational diversity and practical feasibility in automated clustering scenarios. Table \ref{tab:hyperparameters} summarizes the %specific
search spaces and %fixed
parameters employed in our simulation study. For all configurations, parameters not explicitly mentioned in this table were maintained at their default values as defined in the \textit{scikit-learn} framework. The applied strategy ensures a reasonable balance between optimization quality and computational complexity for each clustering algorithm considered.

\subsection{Cluster Validity Indices (CVIs)}

After running the clustering algorithms presented in Section \ref{subsec:clustalg} on our dataset repository, evaluating their effectiveness using CVIs is imperative. In the situation when the true object labels (i.e. the ground truth) are unknown, these indices can be used: (1) to identify the optimal number of clusters in a given dataset, (2) to estimate the quality of an obtained clustering, or (3) to determine the best clustering algorithm for the data at hand. Here, we considered four popular CVIs, including Silhouette \cite{Rousseeuw87}, Calinski-Harabasz (CH) \cite{Calinski74}, Davies–Bouldin (DB) \cite{davies79}, and Dunn \cite{dunn73}.

We used the Silhouette and Dunn indices, following the approach of Brock et al. \cite{brock08}, who proposed to employ them to recommend the most suitable clustering method for the data at hand. Furthermore, we evaluated the performance of the CH and DB indices, as CH is robust in noisy data and skewed distributions, and DB is known for its efficacy in the presence of cluster overlap \cite{arbelaitz13, liu10}.

To compare these indices with ClustRecNet when the true cluster labels are known (as is always the case for our synthetic data), we used ARI, an enhanced version of the Rand Index similarity measure, that accounts for chance correlations and ranges from -1 to 1 \cite{Hubert85}. ARI enables evaluation of the concordance between predetermined classifications and those derived by the clustering algorithms \cite{Chacon23}. Perfect congruence between two clusterings yields an ARI value of 1, whereas the ARI value of -1 indicates that the compared clustering solutions are completely different.

Consider a dataset comprising \(n\) objects partitioned in two distinct ways: \(X\) with \(r\) clusters and \(Y\) with \(s\) clusters. The relationship between these partitions can be represented by a contingency table \([n_{ij}]\), where each cell \(n_{ij}\) indicates the cardinality of the intersection between \(X_i\) and \(Y_j\). ARI is defined as follows:
\begin{equation}\label{eq:ari}
ARI = \frac{\sum_{ij} \binom{n_{ij}}{2} - \left[ \sum_i \binom{a_i}{2} \sum_j \binom{b_j}{2} \right] / \binom{n}{2}}{\frac{1}{2}\left[ \sum_i \binom{a_i}{2} + \sum_j \binom{b_j}{2} \right] - \left[ \sum_i \binom{a_i}{2} \sum_j \binom{b_j}{2} \right] / \binom{n}{2}},
\end{equation}
where \(n_{ij}\) is the number of elements that are common to both cluster \(X_i\) from partition \(X\) and cluster \(Y_j\) from partition \(Y\), and \(a_i\) and \(b_j\) refer, respectively, to the number of elements in clusters \(X_i\) and \(Y_j\).

In our experimental settings, following the guidelines established by Steinley \cite{steinley2004properties}, an ARI value greater than or equal to 0.8 indicated a superior clustering performance (i.e. when the obtained clustering solution demonstrated high concordance with the ground-truth labels). Consequently, for each dataset from our data repository, we used a binary vector of length 10 (as we considered 10 clustering algorithms), in which the value of 1 indicated that the corresponding clustering algorithm provided an ARI value greater than or equal to 0.8, and the value of 0, otherwise. This binary vector constitutes pivotal information for the model, enabling the recommendation system to learn and suggest the most appropriate clustering algorithm(s) for a given dataset. To provide additional transparency on this labeling scheme, we report its empirical distribution in Section \ref{subsec:snyth_exp}.

Silhouette (Sil) is a well-established index for evaluating clustering quality. It is computed using the mean intra-cluster distance and the mean nearest-cluster distance \cite{Rousseeuw87}. The calculation of Sil for cluster \(k\) is carried out using Eq. \ref{eq:sil_k}:
\begin{equation}\label{eq:sil_k}
Sil(k) = \frac{1}{N_k} \left[ \sum_{i=1}^{N_k} \frac{b(i) - a(i)}{\max(a(i), b(i))} \right].
\end{equation}
Here, \(N_k\) is the number of objects in cluster \(k\), \(a(i)\) is the average intra-cluster distance between object \(i\) from cluster \(k\) and all other objects in cluster \(k\), and \(b(i)\) is the smallest average inter-cluster distance between object \(i\) from cluster \(k\) and all objects in the cluster that is the nearest to \(k\).

The Silhouette score corresponding to a given clustering solution comprising $K$ clusters is defined as follows: 
\begin{equation}\label{eq:sil}
Sil(K) =  \sum_{k=1}^{K} \frac{SH(k)}{K}.
\end{equation}
The Silhouette score ranges from -1 to 1. Higher values indicate better-defined clusters. 

The Calinski-Harabasz index \cite{Calinski74} is another popular criterion used to assess clustering solutions. CH is a weighted ratio of the between-cluster and within-cluster dispersions as defined by Eq. \ref{eq:cal}:
\begin{equation}\label{eq:cal}
CH = \frac{tr(B_K)}{tr(W_K)} \times \frac{N - K}{K - 1},
\end{equation}
where \(tr(B_K)\) is the trace of the between-cluster scatter matrix, \(tr(W_K)\) is the trace of the within-cluster scatter matrix, \(N\) is the number of objects, and \(K\) is the number of clusters. The between-cluster scatter matrix \(B_K\) captures the dispersion of cluster centroids around the global centroid, whereas the within-cluster scatter matrix \(W_K\) accounts for the dispersion of data points around their respective cluster centroids. Higher CH values correspond to better-defined clusters. 

Another popular index used in our experiments is the Davies–Bouldin score \cite{davies79}. For a dataset partitioned into \(K\) clusters, DB is defined as follows:
\begin{equation}\label{eq:db}
DB = \frac{1}{K} \sum_{i=1}^K \max_{j \neq i} \left( \frac{\sigma_i + \sigma_j}{d(c_i, c_j)} \right),
\end{equation}
where $\sigma_i$ (or $\sigma_j$) is the average distance between the objects in cluster \(i\) (or \(j\)) and its centroid $c_i$ (or $c_j$), and $d(c_i, c_j)$ is the distance between the centroids of clusters \(i\) and \(j\). Lower DB values indicate a superior clustering performance. 

Finally, we also investigated how the Dunn index \cite{dunn73} can be used for recommending suitable clustering algorithms for a given dataset. It can be calculated using Eq. \ref{eq:dunn}:
\begin{equation}\label{eq:dunn}
Dunn = \frac{\min\limits_{1 \leq i < j \leq K} \delta(i, j)}{\max\limits_{1 \leq k \leq K} \Delta(k)},
\end{equation}
where \(K\) is the number of clusters, \(\delta(i, j)\) is the inter-cluster distance between clusters \(i\) and \(j\), and \( \Delta(k) \) is the diameter of cluster \(k\) (i.e. the maximum distance between objects of cluster $k$). A higher Dunn index indicates a better clustering performance, reflecting well-separated and compact clusters.

\subsection{Deep Model Architecture}
In our experiments, we comprehensively evaluated various neural network architectures to establish a robust baseline and ensure the efficacy of our proposed model. First, we examined feedforward neural networks (FNNs) with various complexities, implementing configurations with 3, 5, and 10 layers, to evaluate their performance on synthetic data from our repository. Subsequently, we explored basic CNN architectures \cite{lecun1998lenet} incorporating 3 and 5 convolutional layers. These preliminary experiments provided critical insights into the strengths and limitations of various network architectures, supplying necessary information for our hybrid approach.

To extract the necessary features for clustering algorithm recommendation \cite{Rice1976alg}, we propose a novel DL architecture, integrating a CNN \cite{lecun1998lenet}, two ResNets \cite{he2016resnet}, and an attention mechanism \cite{Vaswani16}. Such a hybrid model is engineered to harness the inherent advantages of each component, enhancing feature extraction, addressing the vanishing gradient issue, and highlighting crucial features within the input data. The use of a CNN-based architecture for tabular data in ClustRecNet is founded on the principle of density-based feature extraction. While tabular data columns do not possess the spatial continuity of pixels, convolutional layers serve as local filters that scan the feature-object space to identify micro-clusters and density variations. This is complemented by the ResNet blocks, which allow the model to learn the hierarchical structure of these density patterns, and the Attention mechanism, which reconciles the arbitrary order of rows and columns by modeling global dependencies regardless of spatial distance. Unlike Gradient Boosting models, which require meta-feature vectors, this differentiable pipeline allows the network to learn a direct mapping from raw data distributions to clustering algorithm performance.

\begin{figure}[t]
  \centering 
  \includegraphics[width=\linewidth]{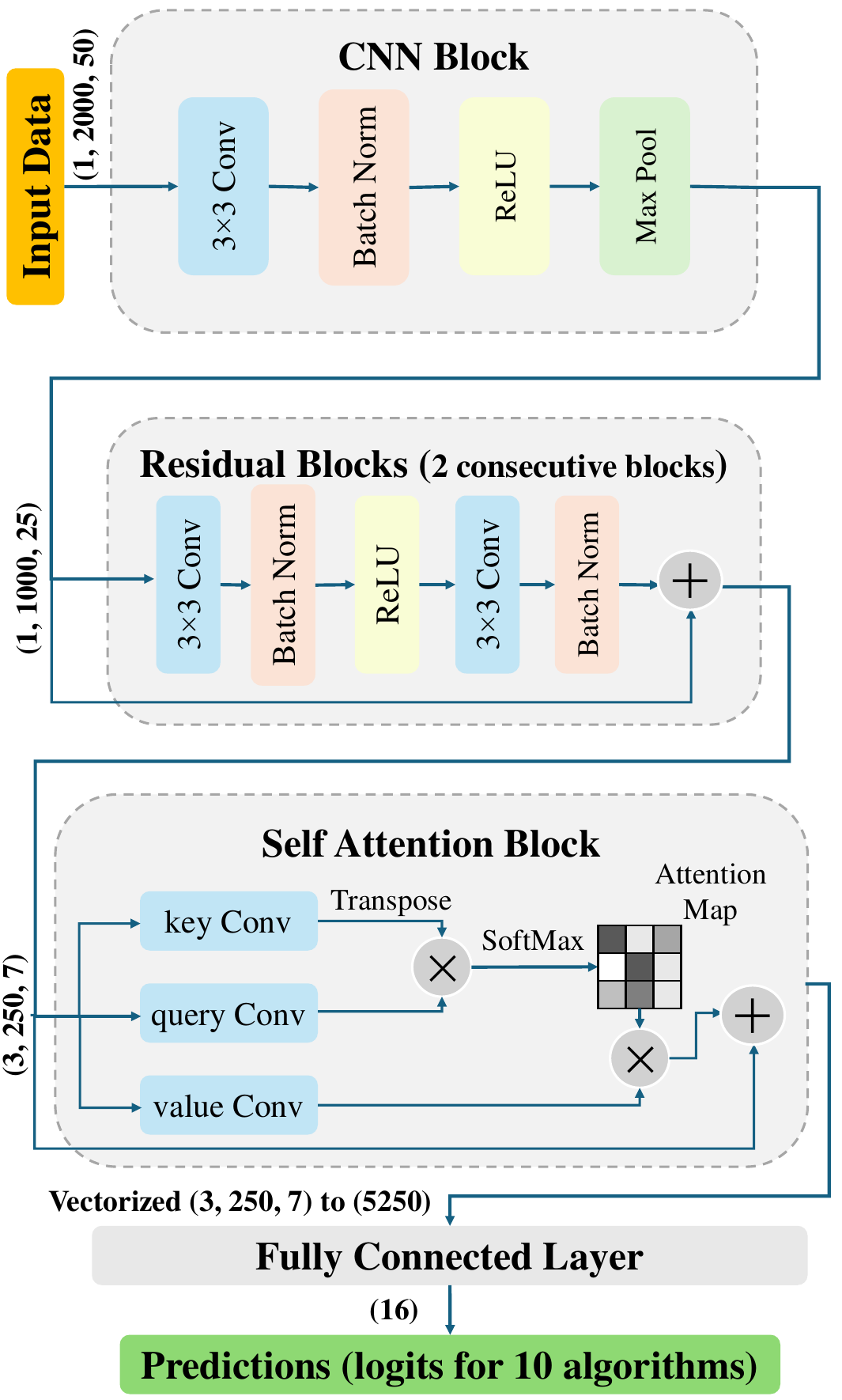}
  \caption{Schematic representation of the proposed ClustRecNet model architecture.}
  \label{fig:modelarc}
\end{figure}

The selected architecture comprises four principal blocks discussed below. Its schematic representation is shown in Fig. \ref{fig:modelarc}.

It starts with a convolutional layer, followed by batch normalization and max pooling. Integrating a 2D convolutional layer, batch normalization, and max pooling is a standard and effective way of organizing an initial processing block in contemporary CNN architectures. After the convolutional layer, the architecture incorporates two consecutive residual blocks. Each of them can be characterized by the following equation:

\begin{equation}\label{eq4}
\mathbf{Y} = F(\mathbf{X}, \mathbf{W}, \mathbf{b}) + \mathbf{X},
\end{equation}

\noindent where $\mathbf{X}$ is the input to the layer, $F(\mathbf{X}, \mathbf{W})$ is the residual mapping to be learned, $\mathbf{W}$ is the set of weight tensors in the residual block, and $\mathbf{b}$ is the bias term. Each residual block comprises two convolutional layers with batch normalization, employing the ReLU activation function. The residual connection, implemented through a shortcut, performs identity mapping or convolution for dimension matching, thereby bypassing the main convolutional layers of the residual block. 

After the residual blocks, we implement a self-attention mechanism inspired by the transformer architecture from Vaswani et al. \cite{Vaswani16}. The attention block can be described by the following equation:
\begin{equation}\label{eq5}
Attention(\mathbf{Q}, \mathbf{K}, \mathbf{V}, \mathbf{X}) = softmax \left( \frac{\mathbf{Q}\mathbf{K}^T}{\sqrt{d_k}} \right) \mathbf{V} + \mathbf{X},
\end{equation}
where $\mathbf{Q}$, $\mathbf{K}$, and $\mathbf{V}$ are learned linear projections of the input features. The attention block uses the learned query, key, and value projections to compute a weighted sum of the input features, enabling the model to capture long-range dependencies within the feature maps and complement the local processing capabilities of the convolutional layers. The output from the attention block is subsequently flattened and processed through two fully connected layers, reaching the final layer that produces logits corresponding to the number of classes in the classification task. 

Therefore, the proposed network architecture is designed to integrate local spatial pattern recognition (via CNNs), stable and hierarchical feature propagation (via residual connections), and global context modeling (via attention), offering a unified solution for learning discriminative representations from raw tabular data. As we will see in the next sections, an effective combination of these components, ClustRecNet, delivers quality performance on complex clustering classification tasks. 

\subsection{Implementation Details}

The described network architecture was chosen after extensive experimentation with multiple network variants, as it provided the best overall performance across our synthetic data repository.
Moreover, the ablation study conducted on real-world data (see Section \ref{subsec:realdata}) showed that all components of the proposed network architecture are essential for generalizing across diverse clustering scenarios. 

Unlike traditional image-based CNNs, our model operates directly on tabular data encoded as 2D arrays, where rows correspond to individual data points (i.e. objects) and columns to features. We treat each dataset as a single-channel "image" of shape  \((1, N, D)\), where \(N\) is the number of objects (padded to a fixed height), and \(D\) is the number of features. This tabular treatment allows the convolutional network to learn local feature patterns across both object and feature dimensions. 

The network starts with a 2D convolutional layer that takes a single-channel input and produces a single-channel output. It uses a 3×3 kernel with a stride of 1 and padding of 1 to preserve spatial dimensions, followed by a batch normalization and a 2×2 max pooling operation to reduce resolution and improve generalization. Next, two residual blocks extract hierarchical features. The first residual block expands the number of channels from 1 to 2 and reduces spatial dimensions by using a stride of 2 in the first convolutional layer. The second block further increases the channels from 2 to 3, using a stride of 2 as well. Each residual block contains two 3×3 convolutional layers, followed by batch normalization. If the spatial or channel dimensions change, a shortcut path with a 1×1 convolution and batch normalization ensures proper residual addition.

To capture long-range interactions, we incorporated into the model architecture a self-attention block, which follows the residual layers. Query and key projections reduce the channel dimension using 1×1 convolutions (with a reduction factor of 8), while the value projection maintains the input channels. Attention weights are computed across flattened spatial positions and applied back to the input with a residual connection. The output is flattened and passed through two fully connected layers - the first compresses the representation to 16 neurons, while the second produces predictions over clustering algorithms. The final model was trained using the Adam optimizer (learning rate \(10^{-4}\), weight decay \(7\times10^{-3} \)). The recommendation task is formulated as a multi-label classification problem using the Binary Cross-Entropy (BCE) loss with logits. This formulation, which accommodates the inherent structural ambiguity of unsupervised learning, identifies a curated set of high-performing clustering algorithms without imposing a strict single-winner constraint. The training proceeds by 30 epochs with a batch size of 32, using 10-fold cross-validation.

To ensure full reproducibility of our results, we maintained a deterministic environment by fixing all random seeds and standardizing our software stack. The ClustRecNet framework was implemented using the PyTorch (v2.9.1) deep learning library. Standard data preprocessing, the implementation of traditional clustering algorithms, and the calculation of ARI and other metrics were performed using scikit-learn (v1.7.2) and NumPy (v2.3.5). Synthetic dataset generation was facilitated by the Repliclust (v1.0.0) package. Additional data manipulation and statistical analyses were supported by pandas (v2.3.3) and SciPy (v1.16.3). Training was performed on an NVIDIA A100-SXM4-40GB GPU, while inference benchmarks were conducted on a workstation equipped with an Intel(R) Core i9-13900 CPU (2.00 GHz). The complete source code, including a comprehensive configuration file for all dependencies and hyperparameter settings, is available on our GitHub repository.

\subsection{Train and Validation Convergence}

\begin{figure}[t]
  \centering 
  \includegraphics[width=\linewidth]{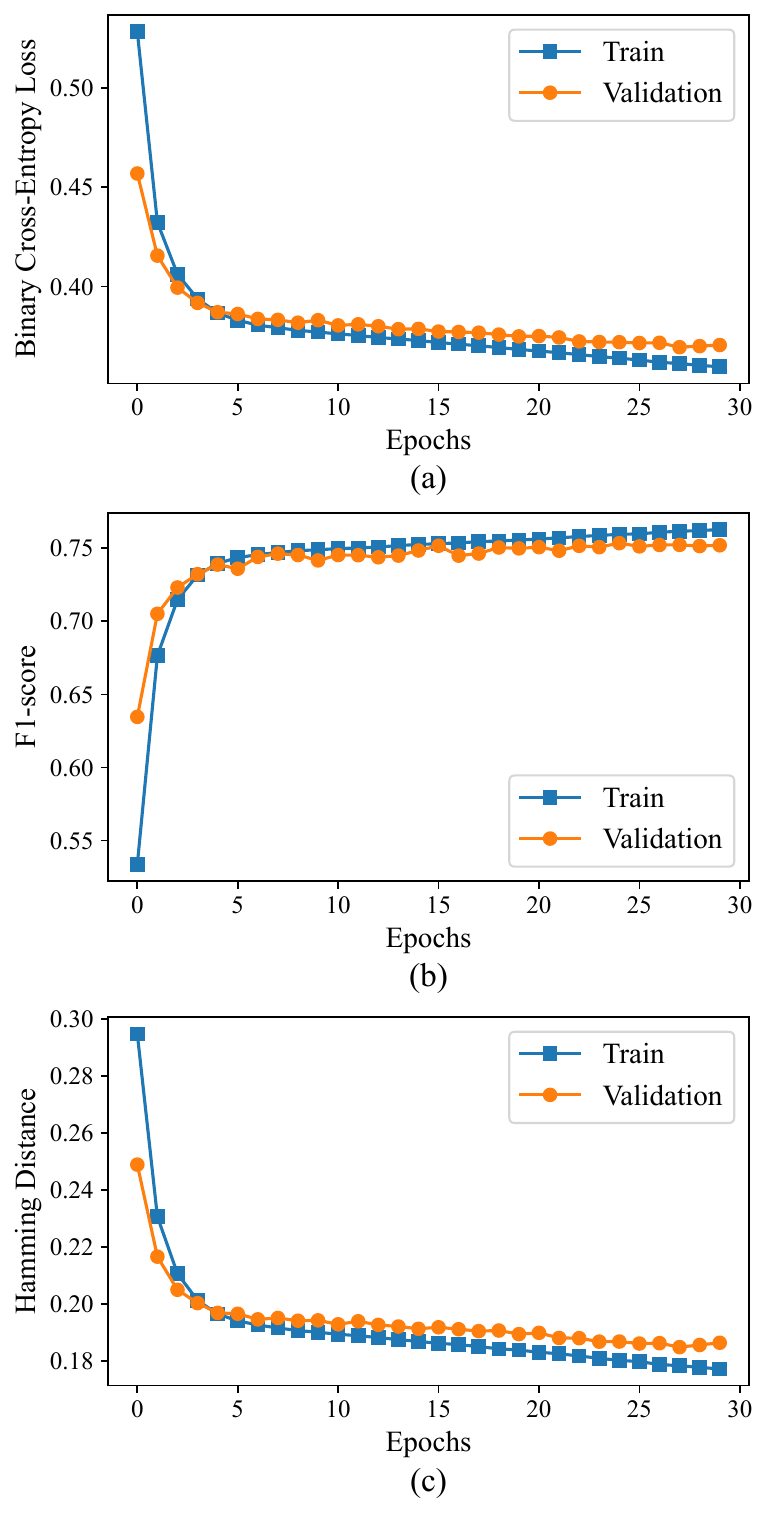}
  \caption{(a) Average loss, (b) average F1-score, and (c) average Hamming distance obtained for training and validation using 10-fold cross-validation across training epochs.}
  \label{fig:ttprog}
\end{figure}

Fig. \ref{fig:ttprog} depicts the average Binary Cross-Entropy (BCE) loss (a), F1-score (b), and Hamming distance (c) results across all folds, obtained using the proposed ClustRecNet model. 

The trends observed in the training and validation loss, F1-score, and Hamming distance results provide insights into the model's convergence behavior and generalization capacity. The BCE loss consistently declines as the number of epochs grows, indicating stable optimization. The training loss drops from 0.528 to 0.360, while validation loss decreases from 0.457 to 0.370. These patterns are reflected in a marked improvement of the F1-score: from 0.534 to 0.763 on training data, and from 0.635 to 0.752 on validation data, thus demonstrating increasing alignment with ground truth clustering labels. At the same time, the Hamming distance — a metric used to quantify binary misclassification — shows a progressive decrease: from 0.295 to 0.177 for training, and from 0.249 to 0.186 for validation. These results correspond to relative improvements of approximately 40\% and 25\%, respectively. A close alignment between the training and validation curves across all metrics highlights strong generalization capacity and stable convergence behavior of ClustRecNet.

\section{Results and Discussion}

\subsection{Experiments with Synthetic Data} \label{subsec:snyth_exp}

In this section, we rigorously evaluate the proposed ClustRecNet recommendation framework, which takes as input an unlabeled dataset, encoded as an $object\times feature$ matrix, and outputs the most suitable algorithm(s) for its clustering, among those described in Section \ref{subsec:clustalg}. During training, 90\% of the 34,000 synthetic datasets considered in our experiments were used for 10-fold cross-validation to ensure robustness and reduce overfitting, while the remaining 10\% were reserved for final model validation. Additionally, weight decay regularization was applied using the Adam optimizer to prevent model overparameterization and enhance its generalization performance.

\begin{figure}[t]
  \centering 
  \includegraphics[width=\linewidth]{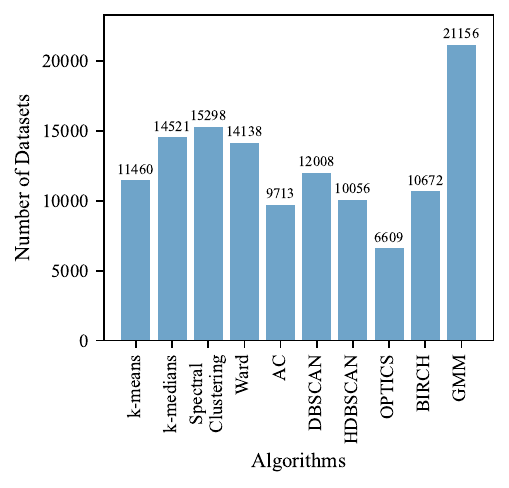}
  \caption{Distribution of algorithm recommendations across our synthetic data repository. The bar chart shows the number of positive recommendations for the 10 clustering algorithms evaluated.}
  \label{fig:ari_recommendations_algorithm}
\end{figure}

Before reporting predictive performance, we examine the empirical structure of the ARI-derived recommendation labels across our whole synthetic dataset repository. As shown in Fig. \ref{fig:ari_recommendations_algorithm}, the number of datasets for which each clustering algorithm achieves $ARI \geq 0.8$ varies significantly, indicating that no single method dominates the experimental space. In our settings, one algorithm, several algorithms, or no algorithms at all could be recommended for a given synthetic dataset. 
Precisely, GMM was recommended for 21,156 (62.2\%) datasets, followed by Spectral Clustering, recommended for 15,298 (45.0\%) datasets, and $k$-medians, recommended for 14,521 (42.7\%) datasets, whereas OPTICS served as a more specialized solution with a total of 6,609 recommendations (19.4\%). 
These results suggest that while GMM provides broad utility across various cluster distributions, density-based methods like OPTICS and HDBSCAN are more sensitive to specific topological requirements, acting as specialized experts rather than general-purpose solvers. The dominance of GMM and Spectral Clustering can be attributed to their flexibility in capturing both linear and non-linear cluster separation boundaries, which were prevalent in our synthetic generator.

% This heterogeneous recommendation distribution, effectively preventing single-algorithm dominance,  
%validates that our synthetic protocol successfully 
% can be accounted for substantial data diversity in our repository.

\begin{figure}[t]
  \centering 
  \includegraphics[width=\linewidth]{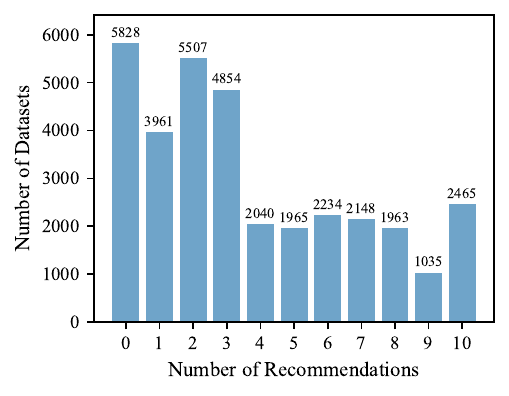}
  \caption{%Distribution of the number of datasets per number of clustering algorithms recommended.
  Distribution of the number of datasets with respect to the number of clustering algorithm recommendations.}
  \label{fig:ari_recommendations_distribution}
\end{figure}

The recommendation cardinality distribution (i.e. the number of datasets that were recommended 0, 1, 2, ..., 10 times) is presented in Fig. \ref{fig:ari_recommendations_distribution}.
%, further characterizes the multi-label nature of the task.
We can observe that 28,172 of 34,000 considered datasets (82.9\%) admit at least one high-performing clustering algorithm. %, with an average of 3.70 recommendations per dataset. 
 A closer examination indicates that approximately 17.1\% (5,828) of the datasets yielded zero recommendations. These cases correspond to scenarios in which conventional clustering methods fail to meet the $ARI \geq 0.8$ criterion, likely due to substantial cluster overlap or elevated noise levels. Furthermore, the predominance of datasets with one to three recommendations (42.1\% of the total) highlights the variability inherent in clustering tasks, where most datasets are associated with a limited subset of well-suited algorithms. The observed range in recommendation cardinality—from single-algorithm suitability to cases where multiple algorithms perform well—supports the formulation of clustering algorithm selection as a multi-label prediction problem, for which binary cross-entropy (BCE) serves as a natural optimization objective.

To ensure a fair comparison of the recommender framework with traditional CVI-based approaches, we conducted a sensitivity analysis for the four selected CVIs (Silhouette, CH, DB, and Dunn) by evaluating their performance across a range of selection thresholds. Table \ref{tab:cvi_thresholds} details how varying these thresholds impacts the corresponding F1-score and Hamming distance evaluation metrics. Based on this grid search, the optimal thresholds were identified as 0.45 for Silhouette, 160 for CH, 0.6 for DB, and 0.6 for Dunn. These optimized configurations were used for the final performance comparison.

\begin{table}[b]
\centering
\begin{tabular}{p{2.2cm}p{1.3cm}|p{2cm}p{1.3cm}}
\toprule
\textbf{Threshold} & \textbf{F1-score} & \textbf{Threshold} & \textbf{F1-score} \\
\midrule
\textit{Silhouette} & & \textit{CH} & \\
0.35 & 0.6987 & 120 & 0.6597 \\
0.40 & 0.7106 & 140 & 0.6634 \\
\textbf{0.45} & \textbf{0.7155} & \textbf{160} & \textbf{0.6679} \\
0.50 & 0.7128 & 180 & 0.6670 \\
0.55 & 0.7017 & 200 & 0.6651 \\
\midrule
\textit{DB} & & \textit{Dunn} & \\
0.4 & 0.6512 & 0.4 & 0.6151 \\
0.5 & 0.6616 & 0.5 & 0.6344 \\
\textbf{0.6} & \textbf{0.6793} & \textbf{0.6} & \textbf{0.6360} \\
0.7 & 0.6730 & 0.7 & 0.6334 \\
0.8 & 0.6545 & 0.8 & 0.6329 \\
\bottomrule
\end{tabular}
\caption{Sensitivity analysis of threshold values for the Silhouette, CH, DB, and Dunn cluster validity indices.}
\label{tab:cvi_thresholds}
\end{table}

\begin{figure*}[h]
  \centering 
  \includegraphics[width=\linewidth]{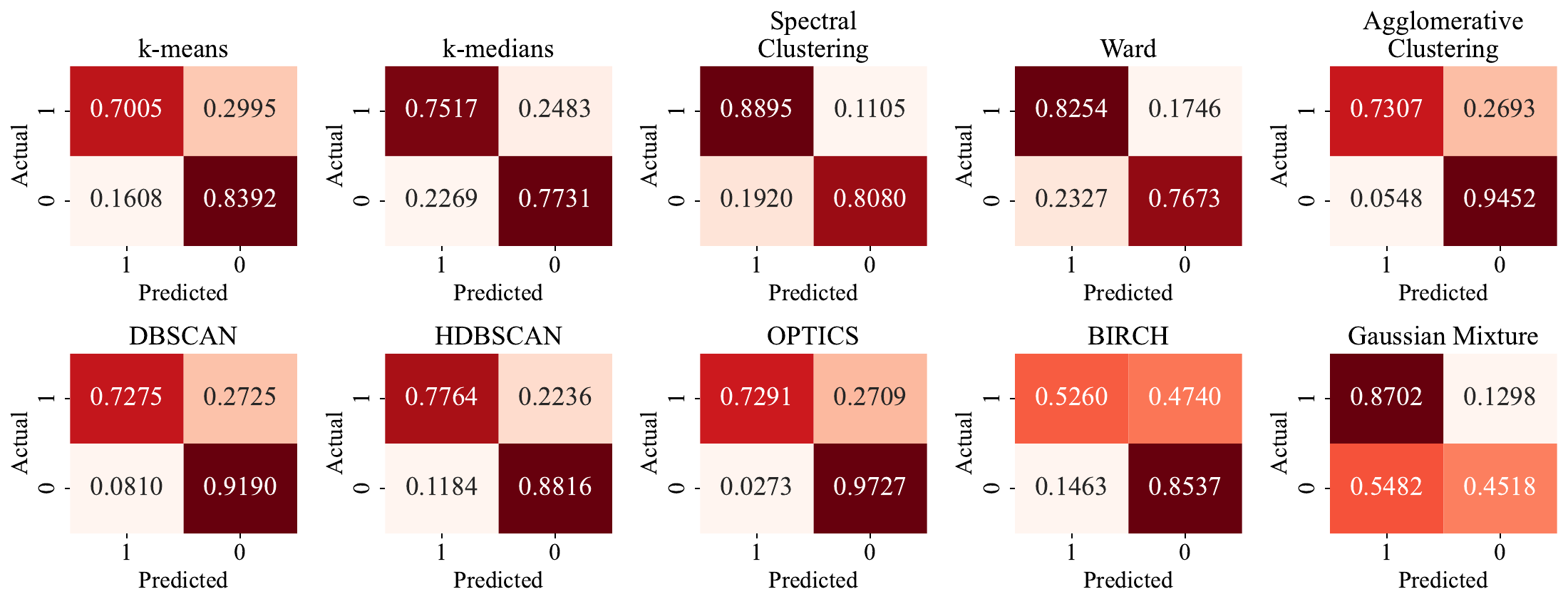}
  \caption{Normalized confusion matrices for the 10 clustering algorithms considered in our study, generated by ClustRecNet on synthetic test data. }
  \label{fig:confusion_matrix}
\end{figure*}

The final comparison results, summarized in Table \ref{tab:synthetic_comparison}, underscore the superior predictive performance of the introduced ClustRecNet model relative to traditional internal validation indices. For each data sample, the optimal algorithm was independently identified by maximizing the Silhouette, CH, or Dunn indices, or by minimizing the DB index. For ClustRecNet, the clustering algorithm corresponding to the highest output logit was selected as the primary recommendation. While ClustRecNet was designed to identify multiple suitable algorithms, this top-1 logit-based selection ensures a direct and fair comparison with traditional CVIs. Using these recommendations, the selected clustering algorithms were executed with hyperparameter configurations optimized via grid search (Section \ref{subsec:hyper}) to compute the reported average ARI values. ClustRecNet achieved a mean F1-score of 0.757 and a mean ARI of 0.878, outperforming the leading CVIs by over 5\% in terms of F1-score (compared to Silhouette) and by 0.497 in terms of ARI (compared to CH). Furthermore, the model reduced the mean Hamming distance to 0.183, which represents a 34.6\% reduction in misclassification compared to the optimized Silhouette baseline (0.280) — thereby confirming its robust capacity to identify optimal clustering structures within diverse synthetic datasets.

\begin{table}[t]
\centering
\small
\caption{Comparison of F1-score, Hamming distance, and ARI results across different methods (ClustRecNet and Silhouette, CH, DB, and Dunn CVIs) on the synthetic test data. The results are reported as means $\pm$ standard deviations. \\}
\begin{tabular}{lccc}
\toprule
\textbf{Method} & \textbf{F1-score} ($\uparrow$) & \textbf{Hamming} ($\downarrow$) & \textbf{ARI} ($\uparrow$) \\
\midrule
ClustRecNet & \textbf{0.757 $\pm$ 0.017} & \textbf{0.183 $\pm$ 0.011} & \textbf{0.878 $\pm$ 0.007} \\
Silhouette     & 0.720 $\pm$ 0.009 & 0.280 $\pm$ 0.009 & 0.365 $\pm$ 0.020 \\
CH             & 0.669 $\pm$ 0.013 & 0.331 $\pm$ 0.013 & 0.381 $\pm$ 0.017 \\
DB             & 0.681 $\pm$ 0.014 & 0.319 $\pm$ 0.014 & 0.344 $\pm$ 0.016 \\
Dunn           & 0.637 $\pm$ 0.011 & 0.363 $\pm$ 0.011 & 0.280 $\pm$ 0.017 \\
\bottomrule
\end{tabular}
\label{tab:synthetic_comparison}
\end{table}

Moreover, we carried out the Wilcoxon signed-rank test to assess whether the improvements over traditional CVIs, achieved by ClustRecNet, were statistically significant. The test results confirm the superiority of our model, with the obtained $p$-values < 0.005 across all three evaluation metrics, strongly rejecting the null hypothesis of equal performance.

To provide a more detailed breakdown of prediction outcomes across 10 clustering algorithms considered in our study, we present their confusion matrices (see Fig. \ref{fig:confusion_matrix}). All reported values are normalized, making them directly comparable across clustering algorithms. As shown, our proposed model consistently identifies suitable clustering algorithms with high precision and recall, particularly using Spectral Clustering, Agglomerative Clustering, DBSCAN, HDBSCAN, and OPTICS, each exhibiting an accuracy over 80\%. Analysis of these matrices reveals distinct trade-offs between sensitivity and specificity. As can be observed, Spectral Clustering and GMM belong to the high-sensitivity group, achieving recall rates of $88.95\%$ and $87.02\%$, respectively. While these models are highly effective at capturing positive instances, they exhibit lower true negative rates, with GMM achieving the lowest specificity of $45.18\%$. In contrast, density-based approaches such as DBSCAN, HDBSCAN, and OPTICS demonstrate high specificity. Notably, OPTICS achieves a near-perfect specificity of $97.27\%$, albeit at the cost of lower recall for the positive class. Other algorithms, including Ward, $k$-medians, and Agglomerative Clustering, provide a balanced compromise, maintaining recall above $73\%$ while keeping specificity above $75\%$. This diversity in performance characteristics underscores the flexibility of ClustRecNet, allowing for strategic selection of clustering paradigms.

Furthermore, we conducted experiments with various random sampling techniques to mitigate the effect of data imbalance. Despite our efforts, these approaches failed to provide significant result improvements, indicating the need for more sophisticated novel approaches that can effectively handle the nuances of multi-label imbalanced datasets. 

\subsection{Experiments with Real Data}\label{subsec:realdata}

To assess the effectiveness of the proposed recommendation framework in practical settings, we conducted experiments on 20 well-known real-world datasets from the UCI Machine Learning Repository\footnote{https://archive.ics.uci.edu}. These datasets, summarized in Table \ref{tab:dataset_summary}, exhibit a diverse range of dimensional characteristics %including the number of objects, features, and classes,
and encompass a variety of domains and complexities. They include popular Ecoli, Glass, Iris, and Wine quality datasets, used in numerous unsupervised and supervised classification studies.

Since the ground-truth labels are available for all these datasets, we were able to compute ARI for each clustering solution obtained to compare the performance of our recommendation model to those of state-of-the-art AutoML-based clustering methods, including ML2DAC, AutoCluster, and AutoML4Clust (AML4C). We selected ARI as our primary evaluation metric as it is the standard benchmark reported in the existing literature. While other metrics, such as Normalized Mutual Information (NMI), can offer further insights \cite{babu2019nmi}, our focus on ARI ensures a consistent and direct comparison with state-of-the-art methods like ML2DAC, whose official implementations primarily report performance in terms of ARI without providing the underlying cluster labels.

For each real-world dataset reported in Table \ref{tab:dataset_summary},  a multi-stage preprocessing pipeline was executed. Categorical variables were transformed into numerical features via one-hot encoding, whereas missing values were imputed using the feature-wise median. To ensure architectural compatibility with the neural network and maintain computational efficiency, datasets exceeding 2,500 instances were subjected to random subsampling. Subsequently, all features were standardized using z-score normalization. These processed datasets served as the input for the proposed model to determine the most suitable clustering algorithm for a given tabular dataset. The selected algorithm was then executed with default scikit-learn parameters over a predefined range of candidate numbers of clusters. Either the Silhouette or CH index was then used to determine the optimal number of clusters. Such an approach simulates a typical unsupervised clustering scenario, where neither ground-truth labels nor the number of clusters are known at inference time.

\begin{table}[t]
\centering
\caption{Summary of real datasets from the UCI repository, including their application domains, number of objects, features, and classes.}
\resizebox{\columnwidth}{!}{
\begin{tabular}{llccc}
\toprule
\textbf{Dataset} & \textbf{Domain} & \textbf{Objects} & \textbf{Features} & \textbf{Classes} \\
\midrule
Breast tissue      & Medicine       & 106  & 9  & 6 \\
Ecoli              & Biology        & 336  & 7  & 8 \\
Glass              & Forensics      & 214  & 9  & 7 \\
Haberman           & Medicine       & 306  & 3  & 2 \\
Iris               & Biology        & 150  & 4  & 3 \\
Parkinsons         & Medicine       & 195  & 22 & 2 \\
Transfusion        & Medicine       & 748  & 4  & 2 \\
Vehicle            & Transportation & 846  & 18 & 4 \\
Vertebral column   & Medicine       & 310  & 6  & 3 \\
Wine quality red   & Food Science   & 1599 & 11 & 6 \\
Annealing          & Industrial     & 898 & 38 & 5 \\
Banknote auth.       & Finance        & 1372 & 4 & 2 \\
Cervical cancer      & Medicine       & 858 & 32 & 5 \\
Chatfield            & Finance        & 235 & 12 & 2 \\
Diabetes             & Medicine       & 768 & 8 & 2 \\
Seeds                & Agriculture    & 210 & 7 & 3 \\
Steel plates faults  & Manufacturing  & 1941 & 27 & 7 \\
Thyroid disease      & Medicine       & 215 & 5 & 3 \\
Tic-Tac-Toe          & Game Theory    & 958 & 9 & 2 \\
Waveform             & Signals        & 5000 & 40 & 3 \\
\bottomrule
\end{tabular}
}
\label{tab:dataset_summary}
\end{table}

After selecting a clustering algorithm and estimating the number of clusters, we performed a grid search to identify suitable hyperparameters for that specific algorithm. The grid values for each parameter were predefined based on commonly used configurations in the literature to ensure consistency across our evaluation. The selected clustering algorithm was then used with the best configuration found, while the clustering quality was assessed using ARI.

\textbf{Ablation Study.} To isolate the contribution of each architectural component of our model, we performed an ablation study in which one of its main components - either the CNN block, or two ResNet blocks, or the attention mechanism (ATN) - was removed, one at a time. The results, summarized in Table \ref{tab:ablation}, highlight the importance of all three main components working in synergy. For the CH-based variant, the complete integrated model (All) reaches a peak mean ARI of 0.2520 and a median of 0.1794. The exclusion of any single module incurs a performance penalty; most notably, removing the attention mechanism (No ATN) results in a 53.9\% reduction in median ARI, whereas removing the ResNet blocks reduces the mean ARI to 0.2008. Although the \textit{No CNN} configuration maintains a relatively high mean, its median ARI falls to 0.1339, underscoring CNN’s vital role in maintaining performance consistency across diverse datasets. A similar pattern can be observed for the Silhouette-based models: while the \textit{No CNN} variant shows a marginal increase in mean ARI, it suffers a clear degradation in median ARI (falling from 0.1684 to 0.1498). These findings demonstrate that while specific modules contribute to local feature extraction or residual learning, their full interaction ensures the model's robustness and superior clustering accuracy, justifying the inclusion of the full architectural suite.

% To isolate the contribution of each architectural component of our model, we performed an ablation study in which one of its main components - either the CNN block, or two ResNet blocks, or the attention mechanism (ATN) - was removed, one at a time. The results, summarized in Table \ref{tab:ablation}, highlight the importance of all three main components working in synergy. The full CH-based model provided the highest mean ARI of \hl{0.2520} and a median ARI of \hl{0.1794}. In contrast, the average ARI across all ablated models dropped to \hl{0.2066} (for mean) and to \hl{0.1124} (for median), corresponding to losses of \hl{21.97\%} and \hl{59.60\%}, respectively. Compared to the closest CH-based ablation, the improvement in mean was rather modest, i.e. \hl{1.5\%}, but the median gain was \hl{25.36\%}, suggesting a much more stable clustering performance, confirming the cumulative value of all components of the model.

\begin{table}[b] 
\centering
\caption{Results of our ablation study presented in terms of ARI, conducted on the real-world UCI benchmarks from Table \ref{tab:dataset_summary}}.
\begin{tabular}{lcc}
\toprule
\textbf{Components} & \textbf{Median} & \textbf{Mean}\\
\midrule
\textbf{All}: CNN + ResNet + ATN (CH)  & \textbf{0.1794} & \textbf{0.2520} \\
\textbf{All}: CNN + ResNet + ATN (Sil) & 0.1684 & 0.2266 \\
\textbf{No CNN}: ResNet + ATN (CH)  & 0.1339 & 0.2483 \\
\textbf{No CNN}: ResNet + ATN (Sil)  & 0.1498 & 0.2306 \\
\textbf{No ResNet}: CNN + ATN (CH) & 0.0955 &  0.2008 \\
\textbf{No ResNet}: CNN + ATN (Sil) & 0.1233 & 0.1925 \\
\textbf{No ATN}: CNN + ResNet (CH) & 0.0826 & 0.1963 \\
\textbf{No ATN}: CNN + ResNet (Sil) & 0.0890 & 0.1708 \\
\bottomrule
\end{tabular}
\label{tab:ablation}
\end{table}

\begin{table*}[t] 
\centering
\caption{Detailed comparison of clustering algorithm recommendation methods, in terms of ARI, across the real-world UCI benchmarks from Table \ref{tab:dataset_summary}. The best performance in each row is \textbf{bolded}, and the second-best performance is \underline{underlined}.}
\begin{tabular}{lcccccccc}
\toprule
\multirow{2}{*}{\textbf{Dataset}}
& \multicolumn{2}{c}{\textbf{ClustRecNet}}
& \multirow{2}{*}{\textbf{Baseline CNN}}
& \multicolumn{3}{c}{\textbf{AutoCluster}}
& \multirow{2}{*}{\textbf{AML4C}}
& \multirow{2}{*}{\textbf{ML2DAC}} \\
\cmidrule(lr){2-3} \cmidrule(lr){5-7}
& CH & Sil
& \multicolumn{1}{c}{}   % placeholder for Simple CNN
& CH & Sil & DB
& \multicolumn{1}{c}{}   % placeholder for AML4C
& \multicolumn{1}{c}{}    % placeholder for ML2DAC
\\
\midrule
\midrule
Breast tissue & \textbf{0.2682} & \textbf{0.2682} & \textbf{0.2682} & \underline{0.2398} & \underline{0.2398} & \underline{0.2398} & 0.2099 & 0.2099 \\
Ecoli & \textbf{0.5011} & \textbf{0.5011} & \textbf{0.5011} & \textbf{0.5011} & \textbf{0.5011} & \textbf{0.5011} & 0.1535 & \underline{0.3863} \\
Glass & \textbf{0.2453} & \textbf{0.2453} & \textbf{0.2453} & 0.1399 & 0.1399 & 0.1399 & 0.0994 & \underline{0.1472} \\
Haberman & \underline{0.1040} & \underline{0.1040} & 0.0274 & \textbf{0.1556} & 0.0621 & 0.0621 & 0.0323 & 0.0323 \\
Iris & \textbf{0.5681} & \textbf{0.5681} & \textbf{0.5681} & \underline{0.5584} & 0.5438 & 0.5438 & \textbf{0.5681} & \textbf{0.5681} \\
Parkinsons & \underline{0.0514} & \textbf{0.1218} & \underline{0.0514} & 0.0000 & 0.0000 & 0.0000 & 0.0473 & 0.0473 \\
Transfusion & 0.0193 & 0.0193 & 0.0131 & 0.0174 & \underline{0.0304} & 0.0095 & 0.0196 & \textbf{0.0344} \\
Vehicle & 0.1097 & 0.0869 & 0.0878 & 0.0125 & 0.0119 & 0.0088 & \textbf{0.1380} & \underline{0.1333} \\
Vertebral column & \textbf{0.3669} & \textbf{0.3669} & 0.1361 & \textbf{0.3669} & 0.0967 & \underline{0.3369} & 0.1211 & 0.2885 \\
Wine quality red & \textbf{0.0611} & -0.0029 & 0.0157 & 0.0000 & 0.0000 & 0.0000 & \underline{0.0505} & \underline{0.0505} \\
Annealing & 0.0856 & \textbf{0.1247} & 0.0856 & \underline{0.1076} & 0.0224 & 0.0273 & 0.0273 & \underline{0.1076} \\
Banknote authentication & \textbf{0.1837} & \underline{0.1203} & 0.0231 & 0.0870 & 0.0433 & 0.0263 & 0.0263 & 0.0962 \\
Cervical cancer & 0.1021 & \textbf{0.1618} & 0.0854 & 0.0942 & 0.0000 & 0.0349 & 0.0010 & \underline{0.1024} \\
Chatfield & \textbf{0.4884} & \textbf{0.4884} & \textbf{0.4884} & 0.1392 & 0.1848 & 0.2095 & 0.2095 & \underline{0.3942} \\
Diabetes & \underline{0.0775} & \underline{0.0775} & \underline{0.0775} & 0.0713 & 0.0000 & 0.0525 & 0.0000 & \textbf{0.1589} \\
Seeds & \textbf{0.4540} & \textbf{0.4540} & \textbf{0.4540} & 0.0754 & 0.3796 & 0.3796 & \underline{0.4125} & 0.3931 \\
Steel plates faults & \underline{0.1751} & \underline{0.1751} & 0.1452 & \textbf{0.1929} & 0.1048 & 0.1048 & 0.1048 & 0.1573 \\
Thyroid disease & \textbf{0.8427} & 0.4277 & 0.5859 & 0.1278 & \underline{0.6030} & \underline{0.6030} & \underline{0.6030} & 0.1278 \\
Tic-Tac-Toe & \textbf{0.1242} & 0.0125 & 0.0000 & \underline{0.0173} & 0.0000 & 0.0000 & 0.0000 & \underline{0.0173} \\
Waveform & \underline{0.2109} & \underline{0.2109} & \textbf{0.3429} & 0.0435 & 0.0000 & 0.0000 & 0.0000 & 0.0435 \\
\midrule
\textbf{Median} & \textbf{0.1794} & \underline{0.1684} & 0.1119 & 0.1009 & 0.0527 & 0.0573 & 0.0750 & 0.1305 \\
\textbf{Mean} & \textbf{0.2520} & \underline{0.2266} & 0.2101 & 0.1474 & 0.1482 & 0.1640 & 0.1412 & 0.1748 \\
\bottomrule
\end{tabular}
\label{tab:clustering_comparison}
\end{table*}

To conduct a clustering performance comparison on real-world data, we first considered a baseline CNN using the first CNN block of our model architecture. Such a baseline shares all other elements of the pipeline – including training data, cluster estimation, and evaluation setup, allowing for a fair performance comparison. We further benchmarked our model against state-of-the-art AutoML-based clustering recommendation systems. Specifically, we used the official, containerized implementation of ML2DAC (available on GitHub\footnote{https://github.com/tschechlovdev/ml2dac}) – endorsed with a reproducibility badge by SIGMOD – on the same real-world data. Furthermore, we evaluated two built-in meta-learning approaches from the same container: AutoCluster and AML4C, covering a range of meta-learning strategies. The detailed comparative results are reported in Table \ref{tab:clustering_comparison}. Although Dataset2Graph \cite{dilmperis2025dataset2graph} explores a comparable methodology, it was omitted from the experimental evaluation due to the lack of publicly available code and detailed hyperparameter configurations, preventing a fair and transparent performance comparison.

\begin{figure}[t]
    \centering
    \includegraphics[width=0.5\textwidth]{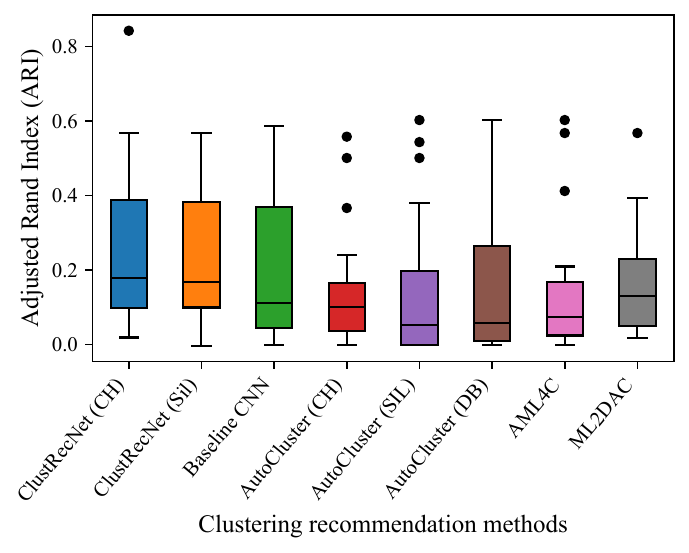}
    \caption{
        Boxplots comparing clustering algorithm recommendation methods across the real-world UCI benchmarks from Table \ref{tab:dataset_summary}.}
        
    \label{fig:ari_boxplot}
\end{figure}

Moreover, we analyzed the ARI distributions using a boxplot diagram (see Fig. \ref{fig:ari_boxplot}) and carried out the Wilcoxon signed-rank test to assess the performance improvement provided by ClustRecNet over the other competing methods (see Table \ref{tab:wilcoxon_comparison}). We found that the ClustRecNet demonstrated statistically significant ARI improvements, with the Wilcoxon test $p$-values $<$ 0.05, over state-of-the-art AutoML-based clustering frameworks, including AutoCluster (CH, Silhouette, and DB variants), AML4C, and ML2DAC, as well as over the baseline CNN. While ClustRecNet yields the highest mean and median ARI across most of the real-world benchmarks, its performance is not unanimously superior; for instance, it ranks as the second-best approach on the Waveform dataset, which suggests that for specific data distributions, a simpler model may offer a more suitable inductive bias than the complex representations learned by ClustRecNet. Furthermore, the performance gap between the CH and Silhouette variants in specific cases, such as the Thyroid and Tic-Tac-Toe datasets, stems from the CVI's difference in estimating the number of clusters. In such instances, if a CVI fails to align with the underlying ground-truth structure, the resulting ARI performance may vary accordingly.

\textbf{Computational Efficiency and Complexity.} To evaluate the practical utility of ClustRecNet, we analyzed its empirical execution time and theoretical time complexity. ClustRecNet demonstrated superior wall-clock performance across our 20 real-world benchmarks. The total pipeline running time for ClustRecNet was 47.39s, which is consistently faster than that of AML4C (75.86s), and orders of magnitude faster than that of ML2DAC (1049.44s) and AutoCluster (290.36s). Notably, the inference of the deep selector itself was computationally negligible, requiring only 0.2 seconds for all 20 datasets, representing less than 1\% of the total runtime. 

Theoretically, the forward pass complexity is determined by the hybrid architecture of the model. For a padded input grid of $R_{\max} \times C_{\max}$, the convolutional and fully connected layers scale linearly, $O(R_{\max}C_{\max})$. The dominant term is the self-attention mechanism, which scales quadratically with the flattened sequence length $S$ after downsampling. For a maximum input grid of $2000 \times 50$, this results in a sequence length of $S = \lceil 2000/8 \rceil \times \lceil 50/8 \rceil = 1750$, ensuring that the deterministic $O(S^2)$ complexity remains computationally tractable for real-time recommendations.

\begin{table}[h]
\centering
\caption{Wilcoxon signed-rank test $p$-values reflecting the performance improvement provided by the proposed ClustRecNet model (CH variant) over the basic CNN and state-of-the-art AutoML approaches; the test was conducted on the real-world UCI benchmarks from Table \ref{tab:dataset_summary}.}
\begin{tabular}{lc}
\toprule 
\textbf{Comparison} & $p$-value \\
\midrule
ClustRecNet (CH) vs Baseline CNN       & 0.0131 \\
ClustRecNet (CH) vs AutoCluster (CH)   & 0.0022 \\
ClustRecNet (CH) vs AutoCluster (Sil)  & 0.0001 \\
ClustRecNet (CH) vs AutoCluster (DB)   & 0.0001 \\
ClustRecNet (CH) vs AML4C              & 0.0001 \\
ClustRecNet (CH) vs ML2DAC             & 0.0045 \\
\bottomrule
\end{tabular}
\label{tab:wilcoxon_comparison}
\end{table}

\textbf{Permutation Invariance.} Furthermore, to demonstrate the robustness of our model, we conducted a comprehensive permutation invariance analysis across {20} real-world benchmark datasets, evaluating the impact of shuffling both row (objects) and column (features) orders of the input matrix. The aggregated results reveal that the model maintains high structural stability; notably, for half of the tested datasets, the model exhibited perfect invariance. Furthermore, the performance metrics remained highly consistent across permutations. Specifically, the mean ARI for row permutations was $0.2220 \pm 0.0149$, while column permutations yielded a mean ARI of $0.2257 \pm 0.0218$. Compared to the ClustRecNet mean ARI of $0.2520$ (calculated on the original input order), these minor fluctuations represent a negligible relative deviation ($<13.5\%$). These empirical findings confirm that through intensive row-and-column shuffling during the training phase, the model achieves empirical permutation equivariance.

While the proposed ClustRecNet model demonstrated strong performance and stability, it also has a few limitations. First, although the model generalizes well to real-world datasets, its predictive accuracy depends on the diversity and coverage of synthetic data used for training. Expanding the training set to better reflect edge-case scenarios, such as highly imbalanced clusters or very noisy data that includes outliers, could further improve its robustness. Second, our current pipeline selects the number of clusters using an internal cluster validity index (CH or Silhouette in our experiments), while all of them are known to be imperfect \cite{arbelaitz13, rykov2024inertia}. Incorporating a learned cluster count estimator or an ensemble of estimators may offer more reliable performance in the future. Third, the model assumes a tabular input representation; extending it to accommodate graph-based or time-series input data could broaden its applicability.

\section{Conclusion}

We developed ClustRecNet – a novel recommendation framework to identify the most suitable clustering algorithm(s) for a given tabular dataset, thus proving that deep learning can be effectively used to tackle this challenging issue. First, we generated a diverse synthetic dataset repository, employing well-established data generation strategies used in clustering. Second, we carried out 10 popular clustering algorithms on each dataset from this repository, assessing the quality of the obtained clustering solutions. The suitability of clustering algorithms for each dataset from our repository was evaluated using ARI, and encoded as a binary vector convenient for multi-label classification tasks. Third, we designed and tested a DL-based hybrid network architecture with a CNN block, two residual blocks, and an attention mechanism. Such an architecture allowed us to extract both low-level and high-level data features, with the attention mechanism highlighting critical features. 

The proposed recommendation framework was evaluated across a variety of synthetic and real-world benchmarks, consistently outperforming conventional cluster validity indices such as Silhouette, Calinski-Harabasz, Davies-Bouldin, and Dunn as well as state-of-the-art AutoML approaches such as ML2DAC, AutoCluster, and AutoML4Clust, thus offering an enhanced practical solution for unsupervised learning tasks. 

Future research could build upon this framework in several directions. First, integrating into the model pipeline dimensionality reduction techniques such as principal component analysis or autoencoders could further optimize the system's performance in high-dimensional spaces, particularly in scenarios where the feature set is large or sparsely populated. Additionally, exploring alternative feature extraction methods such as graph neural networks may enhance the model's ability to capture complex relationships within data, especially for datasets with inherent hierarchical or graph-based structure. Another promising avenue could include extending ClustRecNet to optimize the parameter settings for each clustering algorithm recommended, possibly using reinforcement learning, thus widening its applicability and enhancing precision.

% The final printed size of author photographs is exactly
% 1 inch wide by 1.25 inches tall (25.4 millimeters$\,\times\,$31.75 millimeters/6
% picas$\,\times\,$7.5 picas). Author photos printed in editorials measure 1.59 inches
% wide by 2 inches tall (40 millimeters$\,\times\,$50 millimeters/9.5 picas$\,\times\,$12
% picas).

% Author photographs should be named using the first five characters of the
% pictured author's last name. For example, four author photographs for a
% paper may be named: oppen.ps, moshc.tif, chen.eps, and duran.pdf.

\section*{Acknowledgment}
This work was supported by le Fonds Québécois de la Recherche sur la Nature et les Technologies [grant 371537] and the Natural Sciences and Engineering Research Council of Canada [grant 249644].
% The preferred spelling of the word ``acknowledgment'' in American English is
% without an ``e'' after the ``g.'' Use the singular heading even if you have
% many acknowledgments. Avoid expressions such as ``One of us (S.B.A.) would
% like to thank $\ldots$ .'' Instead, write ``F. A. Author thanks $\ldots$ .'' In most
% cases, sponsor and financial support acknowledgments are placed in the
% unnumbered footnote on the first page, not here. 

\bibliographystyle{plain}
\bibliography{thebibliography}

\EOD

\end{document}